\documentclass[a4paper, 10pt, conference]{ieeeconf}
\IEEEoverridecommandlockouts
\usepackage[hyphens]{url}
\usepackage[pdftex]{hyperref}
\usepackage{array}
\usepackage{tabularx}
\newcolumntype{Y}{>{\raggedright\arraybackslash}X}
\usepackage{cite}
\usepackage{amsmath,amssymb,amsfonts}
\usepackage{algorithmic}
\usepackage{graphicx}
\usepackage{textcomp}
\usepackage{xcolor}
\usepackage[utf8]{inputenc}
 \usepackage{float}
\usepackage{newunicodechar}
\newunicodechar{∞}{$\infty$}
\begin{document}

\title{\LARGE \bf Enhancing the NAO: Extending Capabilities of Legacy Robots for Long-Term Research}
\vspace{-1cm}
\author{Austin Wilson$^{1,2}$, Sahar Kapasi$^{1,3}$, Zane Greene$^{1,2}$, and Alexis E. Block$^{1,2}$
\thanks{$^{1}$SaPHaRI Lab, Human Fusions Institute,}
\thanks{$^{2}$Department of Electrical, Computer, and Systems Engineering,}
        \thanks{$^{3}$Department of Psychological Sciences/Department of Cognitive Science\newline
        Case Western Reserve University, Cleveland, OH, USA
        \newline
        {\tt\small \{amw223, snk83, zdg14, alexis.block\}@case.edu}}%
        \vspace{-1cm}
}

\maketitle

\begin{abstract}
Legacy (unsupported) robotic platforms often lose research utility when manufacturer support ends, preventing integration of modern sensing, speech, and interaction capabilities. We present the \textit{Enhanced NAO}, a revitalized version of Aldebaran's NAO robot featuring upgraded beamforming microphones, RGB-D and thermal cameras, and additional compute resources in a fully self-contained package. This system combines cloud-based and local models for perception and dialogue, while preserving the NAO’s expressive body and behaviors. In a pilot user study validating conversational performance, the Enhanced NAO delivered significantly higher conversational quality and elicited stronger user preference compared to the \textit{NAO AI Edition}, without increasing response latency. The added visual and thermal sensing modalities established a foundation for future perception-driven interaction. Beyond this implementation, our framework provides a platform-agnostic strategy for extending the lifespan and research utility of legacy robots, ensuring they remain valuable tools for human-robot interaction.\looseness-1
\end{abstract}
\vspace{-0.05cm}
\section{Introduction}
\vspace{-0.15cm}
The Aldebaran NAO \cite{aldebaran_nao_2025} has been a staple in human–robot interaction (HRI) research for its approachable design, ease of use, and multimodal sensors. With Aldebaran's recent bankruptcy and uncertainty around ongoing support, many labs risk losing a valuable research and education platform \cite{strathearnSobolewska2025nao}. This challenge is not unique to NAO: Pepper was discontinued in 2021 \cite{nussey2021softbank}, Rethink Robotics shut down in 2018, leaving Baxter obsolete \cite{lawrenceCarol2019rethink, rethink2012baxter}, and Willow Garage closed in 2014, ending support for the TurtleBot (first and second generations) and PR2 \cite{garage2011turtlebots, garage2012pr2, willow_garage}. To address this broader problem of platforms becoming unsupported, we present a platform-agnostic framework for sustaining and enhancing obsolete robots, using the NAO as a central example.

Updating obsolete robots is challenging due to closed designs and outdated onboard compute resources. For NAO, prior work has added external computers \cite{mattamalaNaoBackpack2017}, updated the operating system, \cite{bono2024openaccessnaooan}, or integrated cloud services \cite{gestrin2024NaoChat}, but these approaches trade off portability, scope, or robustness.

\begin{figure}
    \centering
    \includegraphics[width=\linewidth]{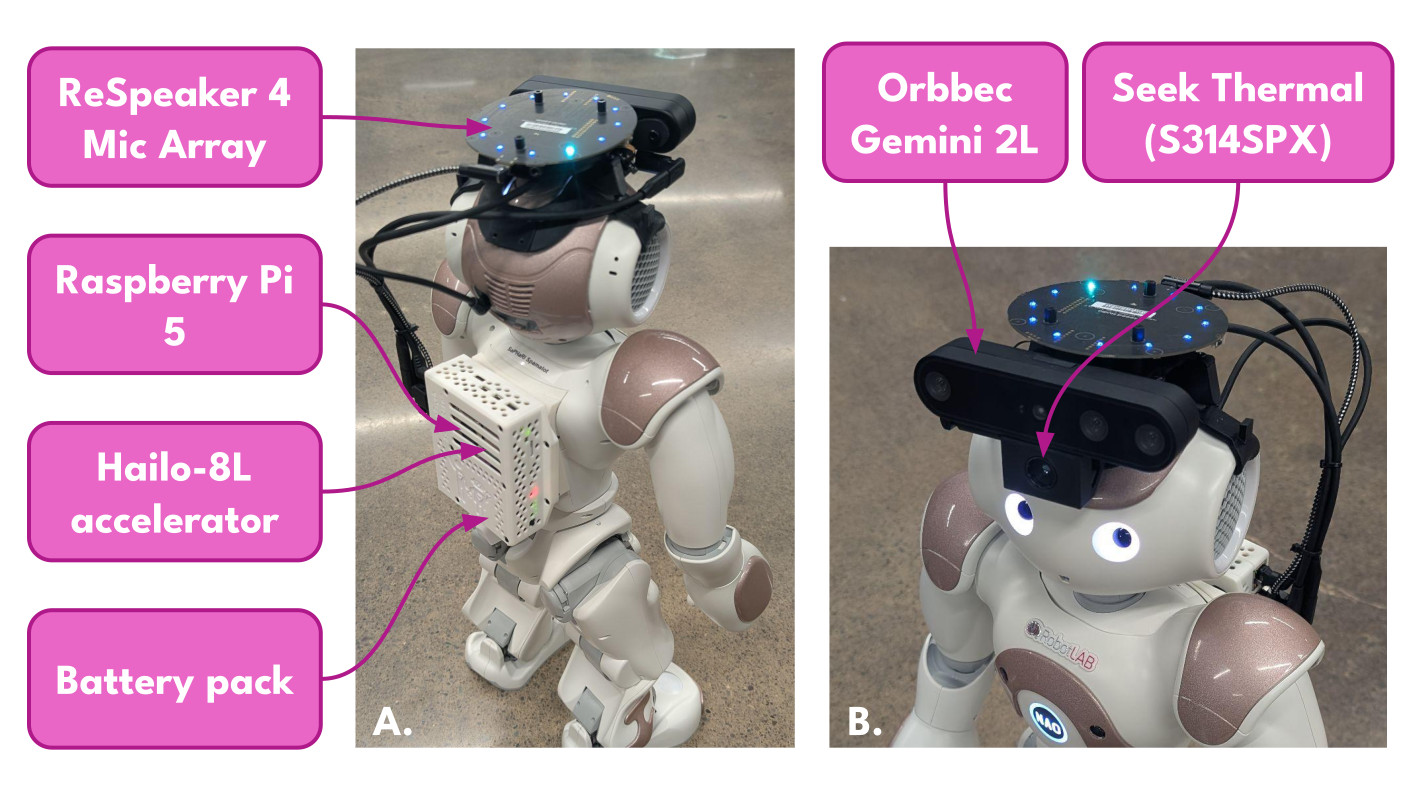}
    \vspace{-0.85cm}
    \caption{Enhanced NAO robot with integrated sensing, computing, and power systems. A.~Back isometric view showing the ReSpeaker 4 Mic Array, Raspberry Pi 5, Hailo-8L, and battery pack. B.~Front view highlighting the Orbbec Gemini 2~L depth camera and Seek Thermal camera (S314SPX).}
    \vspace{-0.5cm}
    \label{fig:NAO_upgraded}
\end{figure}

We address these challenges with an integrated upgrade package (Fig.~\ref{fig:NAO_upgraded}), featuring dual head-mounted cameras for multimodal perception, a Raspberry Pi 5 \cite{rpi5_2025}, a Seeed Studio ReSpeaker 4 Mic Array \cite{respeaker_2023} for robust speech pipelines, and a dedicated battery for peripheral power. This modular design extends sensing, processing, and interaction capabilities without reliance on manufacturer support or tethering.

Here, we contribute: (1)~a complete hardware design, (2)~an integration framework, and (3)~supporting resources available upon request to enable widespread adoption. 
In a pilot validation study, the Enhanced NAO (with the custom upgrade package) demonstrated \textit{richer verbal interaction} (e.g., conversation quality, participant preference) \textit{without added latency} compared to the NAO AI, while \textit{expanded sensing and processing functionalities} establish a practical blueprint for sustaining and enhancing legacy robots.

\vspace{-0.05cm}\section{Related Work}\vspace{-0.15cm}
Researchers have extended the life of legacy research platforms (e.g., NAO \cite{aldebaran_nao_2025}, Baxter \cite{lawrenceCarol2019rethink}, PR2 \cite{garage2012pr2}, TurtleBots \cite{garage2011turtlebots}) through added compute resources \cite{mattamalaNaoBackpack2017, chatzilygeroudis2014naohelmet,bergeon2019raspberry}, updated sensors \cite{chatzilygeroudis2014naohelmet, avalos2017real, da2019control, paul2021object, block2019softness, song2019communication,macias2023map}, modified end-effectors \cite{devine2016real, da2019control, chu2015robotic}, and alternative software stacks \cite{bono2024openaccessnaooan, nao_image_2025, kermorgant2025baxterros2, kumar2024kinematics, wang2016pyride,ramezani2015smooth,alas2016controlling, macias2023map}.

For the NAO, researchers pursued software and hardware strategies\cite{bono2024openaccessnaooan, nao_image_2025, mattamalaNaoBackpack2017, chatzilygeroudis2014naohelmet}. 
Software efforts replaced the proprietary NAOqi head-unit with Ubuntu \cite{nao_image_2025} and ROS2 \cite{bono2024openaccessnaooan}, but these require flashing new firmware\cite{nao_image_2025} and are limited by the NAO's small eMMC, requiring a permanently attached USB stick for operation \cite{nao_image_2025}. Hardware approaches preserved the factory firmware but added external compute and peripherals via short, wired connections \cite{mattamalaNaoBackpack2017, chatzilygeroudis2014naohelmet} , enabling capabilities not supported by the stock head unit.

Similar modifications exist for other humanoid and non-humanoid robots. Prior software upgrades have replaced ROS1 with ROS2 for Baxter \cite{kermorgant2025baxterros2}, abstracted low-level ROS2 interfaces for PR2 \cite{wang2016pyride}, and introduced new inverse kinematic libraries for smoother control of both robots \cite{kumar2024kinematics, ramezani2015smooth}. Hardware upgrades have included custom end-effectors \cite{devine2016real, da2019control, chu2015robotic, fitter2016equipping}, clothing and tactile sensors to improve human-robot hugs \cite{block2019softness}, and RGB-D cameras for teleoperation and object detection \cite{avalos2017real, da2019control, paul2021object}. Non-humanoid robotic platforms, such as the early TurtleBots have received comparable software \cite{song2019communication, alas2016controlling} and hardware upgrades \cite{bergeon2019raspberry, macias2023map}, extending their utility beyond their official support.
\vspace{-0.05cm}\section{Materials and Methods}
\label{sec:materials}
\vspace{-0.15cm}
\begin{table*}[htbp]
    \centering
    \begin{tabular}{|c|c|c|c|}
        \hline
        \textbf{Dimension} & \textbf{1 - Low} & \textbf{2 - Medium} & \textbf{3 - High}\\
        \hline
        \textbf{Relevance ($R$)} & Off topic or mostly unrelated & Partially relevant but incomplete & Fully relevant to the prompt and context \\
        \textbf{Specificity ($S$)} & Vague, generic & Includes some surface level detail & Clear and specific\\
        \textbf{Clarity ($C_l$)} & Hard to follow, confusing & Understandable, some minor issues & Concise, clear, and well-structured\\
        \textbf{Coherence ($C_o$)} & No logical flow from previous turn & Mostly logical & Flows naturally and connects to previous turns\\
        \textbf{Appropriateness ($A$)} & Flat tone & Moderately engaging tone & Warm, polite, and engaging tone \\
        \textbf{Timing ($t$)} & Long pause before response & Robot interrupts user & No issues \\
        \hline
    \end{tabular}
    \vspace{-0.25cm}
    \caption{Rubric for evaluating conversation quality across six dimensions, each rated on a 1-3 scale, with higher scores indicating better quality.}
    \vspace{-0.5cm}
\label{tab:response_rubric}
\end{table*}

Prior work demonstrates the potential and limitations of sustaining unsupported robots via piecemeal upgrades. These efforts highlight the need for integrated solutions. To address these limitations, we developed a platform-agnostic, custom hardware and software upgrade framework that introduces new data streams and enhances on-board compute resources for legacy, yet capable, robots. We demonstrate its usability on the NAO v6, a widely used platform in social and physical HRI research with uncertain long-term support. 

\subsection{NAO Robot}
We use the NAO v6 humanoid robot \cite{aldebaran_nao_2025}, a 58~cm-tall, 5.6~kg bipedal platform with 25 degrees of freedom (DOF), powered by an Intel Atom E3845 CPU with 4~GB RAM and a 32~GB eMMC storage running a Gentoo‑based Linux OS. The robot has two head-mounted HD cameras, a torso-mounted inertial measurement unit (IMU), and two ultrasonic sonars, four directional microphones, two speakers, nine capacitive touch sensors (three in the head, three per hand), and four bumper switches (two per foot). The platform supports Ethernet, Wi-Fi, and USB. Its lithium‑ion battery provides approximately 60-90 minutes of operation. 

The NAO is controlled over a socket interface exposed by the proprietary NAOqi framework.
Aldebaran provided a C++ library, Python bindings, and the Choregraphe GUI, though these require a local network connection to communicate with the socket interface. While the NAO v6 supports local code execution, it is restricted to Python 2.7, limiting developer flexibility \cite{bono2024openaccessnaooan}.

RobotLab further markets the NAO AI Edition (hereafter referred to as ``NAO AI'') with added cloud-based conversational features \cite{robotlab_2022}. In contrast, our work builds on the \textit{base} NAO v6 with custom hardware and software upgrades, collectively referred to as the ``Enhanced NAO:''

\subsection{Hardware upgrades}
\subsubsection{Computing and Power System}
A Raspberry Pi 5 serves as the main processor, featuring a Broadcom BCM2712 quad-core 64-bit Arm Cortex-A76 (Armv8) running at 2.4~GHz \cite{rpi5_2025}. The Raspberry Pi 5 provides USB, GPIO pins, and PCIe expansion, supporting a Hailo-8L accelerator for 13~TOPS of onboard vision-based inference \cite{hailo20258l}.\looseness-1

Peripheral connections include USB ports and CSI/DSI interfaces for additional sensors (e.g., Seek Thermal Camera). The system is powered by a 5~V/5~A USB-C battery pack that sustains all peripherals. Mechanical enclosures, including backpack and sensor housings, were 3D printed, with mounts adapted from open-source designs \cite{chatzilygeroudis2014naohelmet} and custom components, as needed.

\subsubsection{Audio System}
Audio capture is provided by the ReSpeaker with an embedded XMOS XVF-3000, offering built-in echo cancellation, noise suppression, and direction-of-arrival estimation \cite{respeaker_2023}.

\subsubsection{Camera System}
The camera sensing suite integrates an Orbbec Gemini 2~L depth camera \cite{orbbec_gemini} and a Seek Thermal camera (S314SPX) \cite{seek_2024}. The Orbbec provides synchronized RGB-D streams with native OpenCV compatibility, while the Seek Thermal delivers thermography data and RGB thermal visualizations. 

\begin{figure}
    \centering
    \includegraphics[width=1\linewidth]{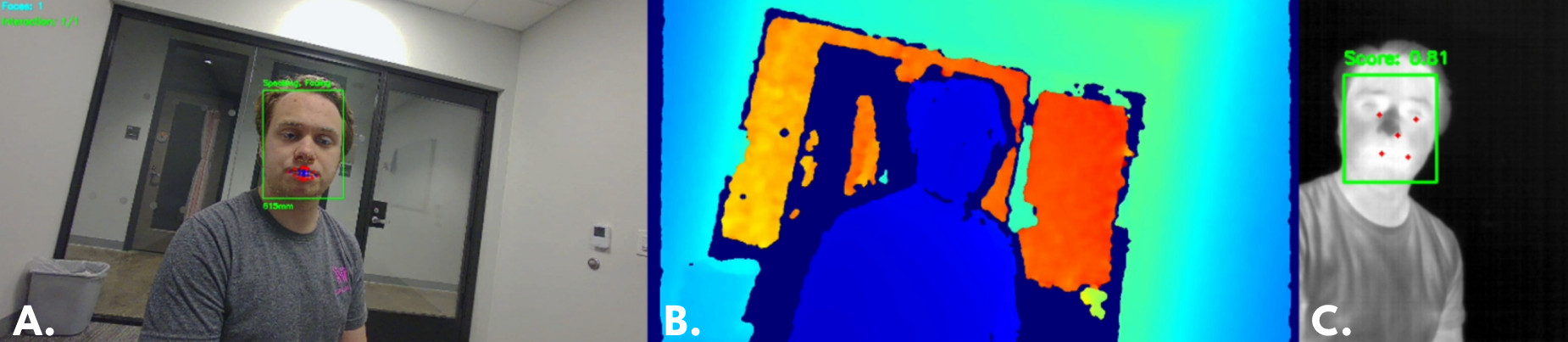}
    \vspace{-0.75cm}\caption{Sample camera frames produced by the enhanced camera system. A)~Processed RGB and B)~depth frames from the Orbbec Gemini 2~L and (C)~processed thermal frame from the Seek Thermal (S314SPX) camera.}
    \vspace{-0.3cm}
    \label{fig:camera_system}
\end{figure}

\subsection{Software upgrades}
The Raspberry Pi 5 runs Raspberry Pi OS  with system nodes implemented in Node.JS, Python, and C++.

\subsubsection{Camera Fusion}
The Raspberry Pi 5 processes RGB, depth, and thermal streams locally. Frames are synchronized and passed through the Hailo-8L for human detection, face region extraction, and landmark identification in thermal and RGB data. This pipeline establishes integration of thermal and visual cues for future interaction logic. Because the Enhanced NAO's upgraded cameras augment rather than replace the existing NAO AI camera system, they were not used in the pilot study. Their integration expands sensing capabilities (adding depth and thermal perception) (Tab.~\ref{tab:feature_comparison}) without interfering with the existing system. However, we include a passive vision pipeline integrating the Orbbec Gemini 2~L camera, Seek Thermal module, and the Hailo-8L for demonstration (Fig.~\ref{fig:camera_system}).

\subsubsection{Verbal Interaction Pipeline}

\begin{figure}
    \centering
    \includegraphics[width=\linewidth]{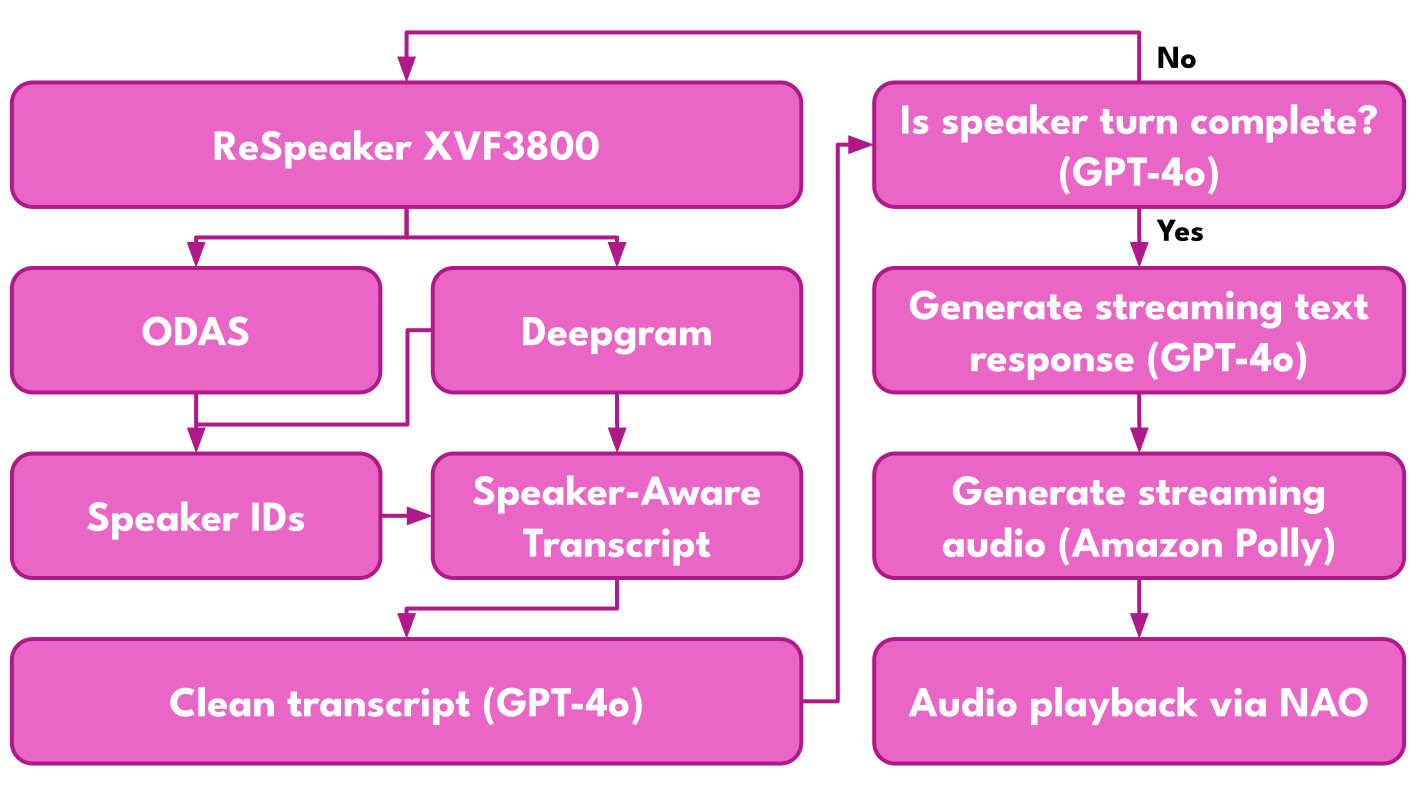}
    \vspace{-0.85cm}
    \caption{Audio processing pipeline, from audio input via the ReSpeaker to a response audio played back through the NAO speakers.}
    \vspace{-0.5cm}
    \label{fig:VerbalPipeline}
\end{figure}

The audio pipeline (Fig.~\ref{fig:VerbalPipeline}) begins with the ReSpeaker, which outputs raw, processed, and echo-canceled streams. These signals are processed by the Open Embedded Audition System (ODAS) \cite{grondin2018lightweightoptimizedsoundsource}, for real-time speaker separation and sound source localization.

Deepgram \cite{deepgram2025transcription} performs speech recognition and completion, bypassing NAO's built-in speech-to-intent engine to enable unconstrained dialogue. Following Deepgram's best practices, speech completion is detected when either the Endpointing feature identifies the end of natural speech in the raw audio stream or when the UtteranceEnd event indicates a sufficient silence gap in the transcript. Once a complete transcript is received, we align the word timings with ODAS outputs to produce speaker-aware transcripts. 

Although Deepgram achieves a low transcription error rate, performance can degrade with low speaker volume, microphone quality, distance, or background noise. Additionally, a complete transcript does not always indicate that a participant has finished speaking. To address these challenges, we created three GPT-4o \cite{openai2024gpt}-based LLM agents with distinct roles: a)~correcting transcription errors in speaker-aware transcripts, b)~detecting turn completion, and c)~generating a contextually appropriate response. We used prompt engineering to structure the system and user prompts for each agent, ensuring accurate task execution and coherent, contextually relevant replies that sustained natural, open-ended dialogue.

To further reduce latency, responses are streamed immediately using Amazon Polly \cite{AmazonPolly_2025}. Because Polly does not natively support streaming, we developed a sentence-segmentation method that sends ordered requests, enabling a ``think and speak'' behavior and reducing the delay between user input and robot output. 

\subsection{Pilot validation study}

\begin{figure}
    \centering
    \vspace{-0.4cm}
    \includegraphics[width=0.75\linewidth]{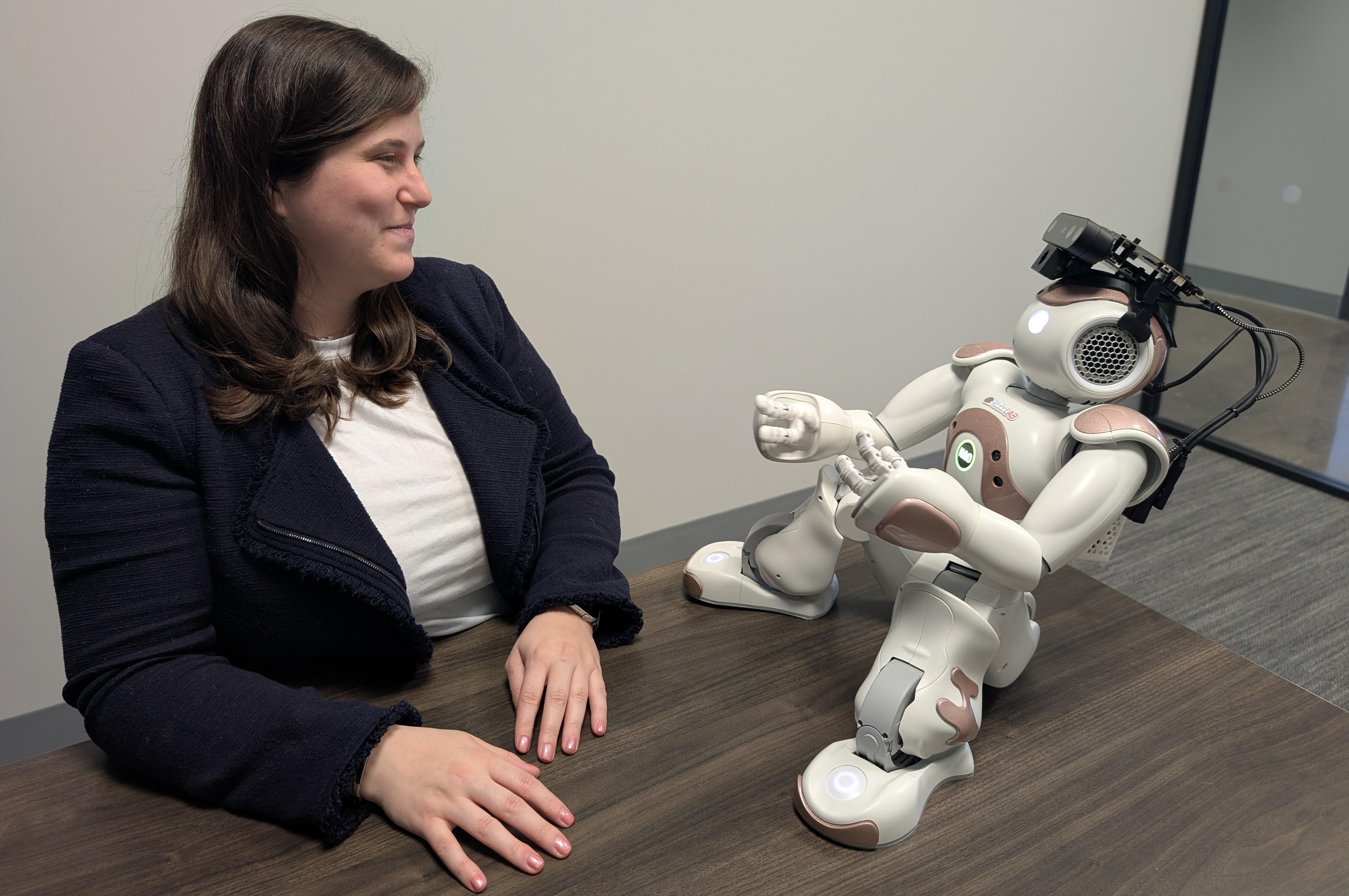}
    \vspace{-0.4cm} \caption{A user engaging in a multi-turn conversation with the Enhanced NAO during the pilot validation study.}
    \vspace{-0.5cm}
    \label{fig:user}
\end{figure}

\begin{figure}
    \centering
    \vspace{-0.5cm}
    \includegraphics[width=\linewidth]{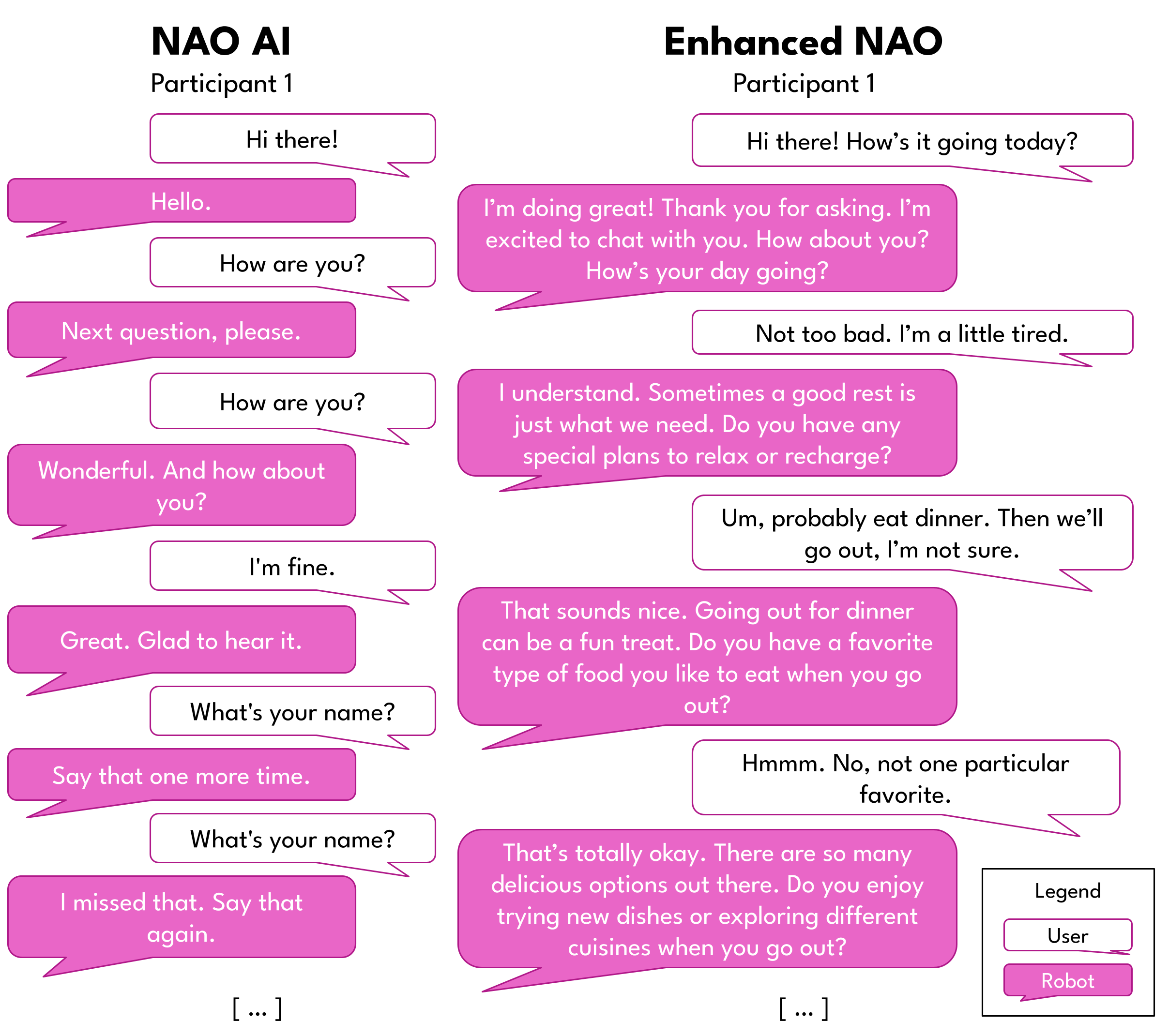}
    \vspace{-0.8cm} \caption{Example of a participant conversation with the NAO AI and the Enhanced NAO.}
    \vspace{-0.5cm}
    \label{fig:exampleconvo}
\end{figure}

We conducted a within-subjects, counter-balanced pilot study with eight participants (four male presenting, four female presenting) to compare interactions with the NAO AI and the Enhanced NAO. The study focused on evaluating conversational capabilities, the only functionality directly replaced and improved by the Enhanced NAO, using equivalent audio inputs and outputs. Each 15-minute session included two conditions, presented in counterbalanced order, with no compensation provided. In each condition, participants held a short conversation ($\geq 4$ \textit{turns}; one turn defined as a user utterance followed by a robot response). As shown in Fig.~\ref{fig:user}, the NAO was placed on a table facing the seated participant, bringing it closer to eye level. To ensure consistency, the first turn was always a user greeting to the robot; subsequent turns were open-ended, with the option to prompt the robot for a joke, if needed. A comparison of example conversations from both conditions is shown in Fig.~\ref{fig:exampleconvo}. After both conversations, each participant was verbally asked which conversation they preferred (e.g., in terms of ease and naturalness of conversation). Their responses were manually recoded by a research assistant. All sessions were video-recorded, manually transcribed, segmented into speaker turns, and timestamped for conversational quality assessment.

\subsection{Quantifying quality of responses}
We assessed conversational quality using a two-level framework adapted from LLM evaluation methods \cite{miller2025evaluating}. The original framework included coherence, accuracy, clarity, relevance, and efficiency; we adapted it for conversational contexts by replacing accuracy with specificity, removing efficiency, and adding appropriateness and timing. This framework allows us to capture objective response quality and subjective user experience. 

At the \textit{turn level}, each robot response was rated on six dimensions (Table~\ref{tab:response_rubric}): \textit{relevance} (addresses user input), \textit{specificity} (detail vs. vagueness), \textit{clarity} (ease of understanding), \textit{coherence} (logical flow), \textit{appropriateness} (politeness and tone), and \textit{timing} (delay before response). 

At the \textit{conversation level}, we computed mean scores for each dimension:
\begin{equation}
\overline{D} = \frac{\sum_{i=1}^{n} d_i}{n}
\end{equation}

\noindent where $\overline{D}$ is the dimension mean, $d_i$ is the rating for the dimension at turn $i$, and $n$ is the number of turns rated. Once we obtain mean values for all six dimensions, we calculate an overall conversational quality score as: 
\begin{equation}
C_q = \frac{\overline{R} + \overline{S} + \overline{C_l} + \overline{C_o} + \overline{A} + \overline{t}}{6}
\end{equation}
\noindent where $C_q$ is the conversation quality score, and $\overline{R}$, $\overline{S}$, $\overline{C_l}$, $\overline{C_o}$, $\overline{A}$, and $\overline{t}$ represent the mean ratings for relevance, specificity, clarity, coherence, appropriateness, and timing, respectively.

We also examined \textit{turn balance}, measuring whether conversations were dominated by the robot or the user. We define this ratio as:  

\begin{equation}
    \label{eq:percent_talking}
    R =  \frac{ W_r}{W_u} 
\end{equation}

\noindent where $W_r$ and $W_u$ denote the number of words spoken by the robot and the user, respectively. Their ratio quantifies relative talkativeness. Using Equation~\ref{eq:percent_talking}, an $R = 1$ indicates perfectly balanced turn lengths, $R > 1$ indicates robot-dominated speech, and $R < 1 $ indicates user-dominated speech. This value is averaged across the entire interaction. Because natural conversation generally involves shared participation, we considered $0.5 < R < 2$ an acceptable range. Values outside this range signal issues in turn-taking, such as overly strict end-of-turn detection (user-dominated conversation) or overly verbose robot responses (robot-dominated conversation), both of which can lead to a less comfortable interaction.

Finally, we tracked \textit{topic shifts} in each conversation, defined as major disruptions to conversational flow. A topic shift was recorded when the robot: 1)~introduced an unrelated subject, 2)~ignored or cut off the user's input, stopping any elaboration, 3)~returned to an earlier subject without linkage, or 4)~closed a topic prematurely. While related to coherence, topic shifts capture disruptive changes in the semantic flow and direction of the conversation, whereas coherence addresses whether consecutive turns of the conversation flow logically. In this way, \textit{most topic shifts} can be \textit{classified as coherence issues}, but \textit{not all coherence issues are topic shifts}.

Two evaluators (co-authors on this paper, who were not involved in the design or implementation of the Enhanced NAO) developed the coding schema before viewing any videos, to minimize expectation bias. All recordings were then coded by a primary researcher, with 20\% independently coded by a secondary researcher to assess reliability. Although evaluators were aware of the condition assignments, all ratings followed standardized criteria to minimize bias. 
\vspace{-0.05cm}\section{Results}
\vspace{-0.15cm}
\begin{table*}[htbp]
\centering
\small
\setlength{\tabcolsep}{7pt}
\renewcommand{\arraystretch}{1.12}
\begin{tabularx}{\textwidth}{|>{\bfseries}p{3.2cm}|Y|Y|}
\hline
Feature & \textbf{NAO AI} \cite{robotlab_2022} & \textbf{Enhanced NAO}\cite{aldebaran_nao_2025} with Hardware/Software Upgrades \\
\hline
Cameras & 

\textit{2 front-facing cameras} (OV5640); 1920$\times$1080 up to 30 FPS; 2560$\times$1920 at 15 FPS \newline Frame rates may be impacted by NAO v6 compute resources.
& 
\textit{In addition to the base NAO camera features:} \newline
Orbbec Gemini 2~L \textit{depth camera}; 1280$\times$800 at 30 FPS
Seek \textit{Thermal camera}; 320$\times$240 at 27 FPS. 

Support for 2 additional CSI Cameras.
\\ \hline
Microphones & 
\textit{Four omnidirectional microphones}; four channels at 48kHz or one channel at 16kHz. Accessible via NAOqi ALAudioDevice in 170ms chunks \cite{aldebaran_nao_2025}. No algorithms to support ASR pipelines.
& 
\textit{Replaces existing NAO microphones} \newline 
Seeed Studio ReSpeaker \textit{4 Mic Array} \cite{respeaker_2023} \newline \textit{Six channels} at 16kHz (4 raw, ASR, AEC) \newline Additional LED indicators.
\\ \hline
Speech (TTS) & 
Converts via NAOqi ALTextToSpeech. Supports pitch, speed, volume, pause, emphasis control, and voice effects. \cite{aldebaran_altexttospeech_2025} & 
Flexible with NAO's default TTS engine, local TTS engines, or cloud-based TTS services. Tested implementation uses Amazon Polly \cite{AmazonPolly_2025}.
\\ \hline
Speech (ASR) & 

Via NAOqi ALSpeechRecognition. \textit{Only recognizes predefined keywords.} Requires a set vocabulary and returns matches with confidence scores. Supports word spotting within longer speech. \cite{aldebaran_asr} &

\textit{Flexible transcription} using a dedicated ASR channel from the ReSpeaker. Tested implementation uses Deepgram to facilitate real-time speech-to-text.\\
\hline 
\textbf{Conversational \newline Capabilities} & 
\textit{Static, intent-driven} system focused on matching predefined intents with spoken keywords that trigger pre-determined responses. TTS engine has limited voices and lacks dynamic range. & 
Supports \textit{open-ended dialogue} with Deepgram \cite{deepgram2025transcription}, uses prompt engineering to tune an LLM (GPT-4o) \cite{openai2024gpt} to generate relevant responses, and supports dynamic TTS voice generation via Amazon Polly \cite{AmazonPolly_2025}.
\\ \hline
\end{tabularx}
\vspace{-0.25cm}
\caption{Comparison of NAO AI capabilities and Enhanced NAO system features.}
\label{tab:feature_comparison}
\vspace{-0.5cm}
\end{table*}

\begin{figure}
    \centering
    \includegraphics[width=1\linewidth]{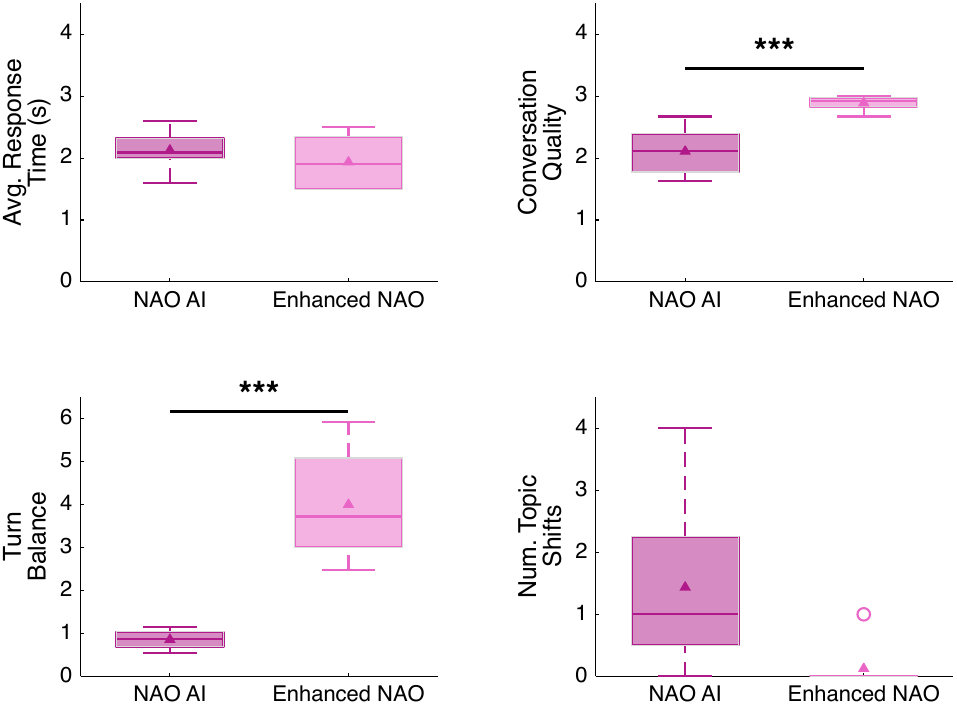}\vspace{-0.25cm}\caption{A comparison of average response times, conversation quality, turn balance, and number of topic shifts between the NAO AI (dark pink) and the Enhanced NAO (light pink). Boxes show the interquartile range (25th-75th percentiles), with the centerline representing the median and triangles marking the mean. Whiskers extend to the most extreme values not considered outliers. Circles indicate outliers. Black lines with asterisks indicate statistically significant differences after Bonferroni correction.}
    \vspace{-0.25cm}
    \label{fig:convo_figs}
\end{figure}

\begin{figure}
    \centering
    \vspace{-0.1cm} \includegraphics[width=1\linewidth]{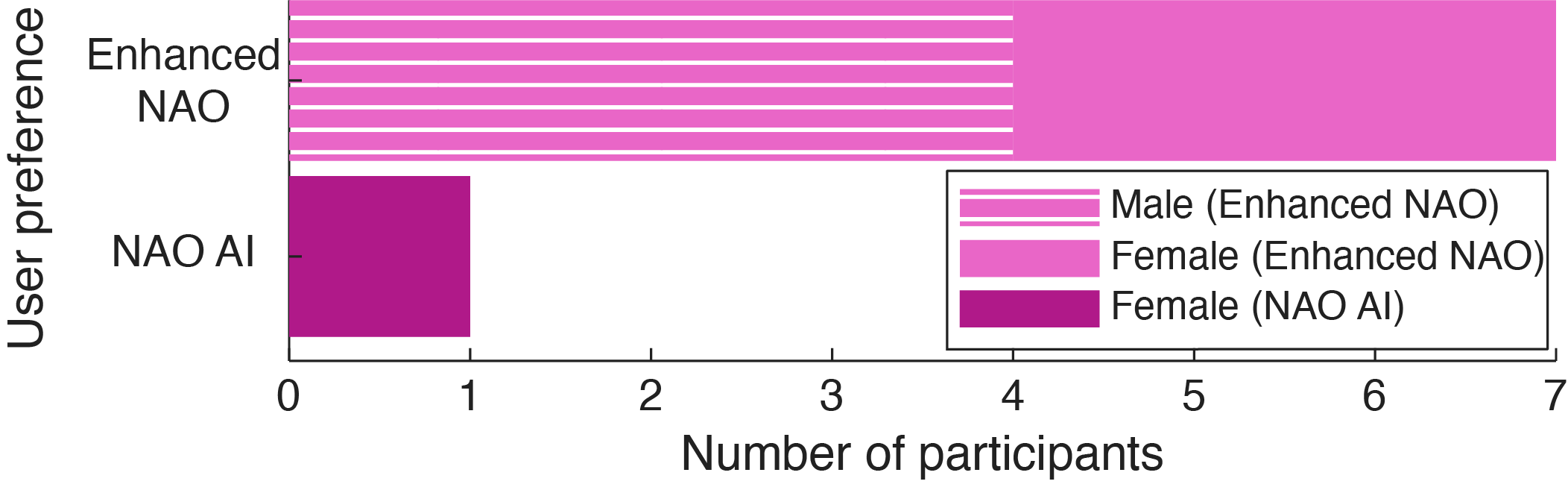}
    \vspace{-0.8cm}\caption{Participant preferences in the pilot study for interactions with the NAO AI (dark pink) versus the Enhanced NAO (light pink). Shading indicates participant gender: stripes indicate male,  solid indicates female.}
    \vspace{-0.5cm}
    \label{fig:user_preference}
\end{figure}

Table~\ref{tab:feature_comparison} summarizes the hardware and software improvements of the Enhanced NAO compared to the NAO AI.
\subsection{Hardware}
 In addition to the NAO's two head-mounted cameras, we integrated a depth and thermal camera, expanding multimodal vision with synchronized RGB-D and thermal streams. The Hailo-8L accelerator allows real-time inference with custom models on these streams, a capability unavailable on the NAO AI. The ReSpeaker handles audio capture, rather than the NAO v6's built-in microphones, bypassing NAOqi's restrictions. NAOqi provides audio only through ALAudioDevice in either 16~kHz or 48~kHz in 170~ms chunks \cite{aldebaran_nao_2025}, without support for beam steering or noise suppression, limiting custom ASR pipelines. 

\subsection{Software}
We restructured the conversational pipeline by replacing the NAO's intent-driven dialogue system with real-time, cloud-based transcription and diarization (Deepgram \cite{deepgram2025transcription}). The Enhanced NAO uses an LLM, rather than fixed templates, for response generation, which enables dynamic, free-flowing conversation. To produce natural-sounding voices, we use Amazon Polly \cite{AmazonPolly_2025} for text-to-speech (TTS) generation. These software modifications are part of the system's functional improvements, which directly contribute to the Enhanced NAO's observed gains in conversational quality and motivated the selection of our evaluation metrics.

\subsection{Conversation}
As described in Sec.~\ref{sec:materials}, all conversations were independently coded by two researchers. Inter-rater reliability \cite{hallgren2012computing}, calculated using a two-way random intraclass correlation coefficient (ICC[2,1]), indicated excellent agreement between coders ($\text{ICC} = 0.9897$).

For each conversation attribute (average response time, conversational quality, turn balance, and number of topic shifts), we performed a paired \textit{t}-test with a Bonferroni correction ($\alpha = 0.0125$) to account for multiple comparisons (Fig.~\ref{fig:convo_figs}). We report effect sizes using Cohen's $d$. 
\textit{Average response time} did not differ significantly between the Enhanced NAO ($M = 1.94, SD = 0.42$) and the NAO AI ($M = 2.13, SD = 0.30$), $t(14) = 1.05, p = 0.3105, d = 0.53$. 
\textit{Conversational quality} was significantly higher for the Enhanced NAO ($M = 2.89, SD = 0.11$) compared to the NAO AI ($M = 2.11, SD = 0.37$), $t(14) = -5.69, p < 0.001, d = -2.85$. 
\textit{Turn balance}, was also significantly higher for the Enhanced NAO ($M = 4.00, SD = 1.19$) than the NAO AI ($M = 0.86, SD = 0.21$), $t(14) = -6.89, p < 0.001, d = -3.44$, indicating that the Enhanced NAO produced substantially more words relative to the user.
Additionally, the Enhanced NAO exhibited fewer \textit{topic shifts} ($M = 0.12, SD = 0.35$) than the NAO AI ($M = 1.44, SD = 1.35$), $t(14) = 2.66, p = 0.0185, d = 1.33$, which approached significance after the alpha correction.

Finally, user preference data (Fig.~\ref{fig:user_preference}) shows that $87.5\%$ of users (seven out of eight) preferred the Enhanced NAO over the NAO AI, citing more natural and engaging conversation.

\vspace{-0.05cm}\section{Discussion}
\vspace{-0.15cm}
The hardware and software upgrades of the Enhanced NAO transformed it into a more capable and sustainable platform by adding sensing modalities, real-time inference, and low-level audio access. These enhancements move beyond the NAO AI's capabilities and limitations to demonstrate how modern compute and perception capabilities can be layered onto legacy hardware. Although we did not benchmark the new camera and microphone system against the NAO's integrated sensors, the additional RGB-D and thermal streams, together with advanced audio processing, provide valuable new data sources and greater developer control. Importantly, the upgrades were delivered in a fully self-contained, untethered package, demonstrating a platform-agnostic path for extending the lifespan and usability of unsupported robots.\looseness-1

These significant hardware upgrades directly improved the \textit{conversational quality} of the Enhanced NAO. Cleaner, lower-latency audio from the ReSpeaker and the LLM-driven conversational pipeline generated more natural, context-aware responses. In contrast, the NAO AI's rigid intent-based models produced repetitive, unnatural, and disruptive replies, reducing flow. \textit{Response times} did not differ significantly between conditions, showing that richer interactions can be delivered without added latency, despite additional sensing, processing, and cloud services. Interestingly, \textit{turn balance} for the Enhanced NAO exceeded our desired range, yet participants still preferred it. This suggests that turn balance may be less critical to user preference than we initially assumed, or that our acceptable range was too conservative. Nevertheless, avoiding overly long or monologic robot responses is still necessary for maintaining enjoyable, user-centered conversation. Additionally, the Enhanced NAO exhibited fewer \textit{topic shifts}, suggesting conversations stayed more coherent and user-driven.\looseness-1

While the primary focus of this work was technical enhancement, we also acknowledge that the visible mechanical additions may have influenced user perception. The NAO's humanoid, friendly appearance supports comfort and approachability, especially among children \cite{shamsuddin2012humanoid, amirova202110, rossi2022using}. Exposing additional hardware elements may make the robot appear more mechanical and potentially reduce approachability. Future adaptations should therefore consider visual and tactile design elements. Prior research on social and affective robots highlights the benefits of softness and warmth for comfort during social-physical interactions \cite{block2019softness, block2017should, block2018emotionally, block2021six, block2023arms, block2021huggiebot}. Inspired by this line of work, future iterations of the Enhanced NAO could incorporate a soft cover or hat over exposed components to preserve the NAO's approachability while maintaining the improved functionality.

\subsection{Limitations}
This work was a pilot validation study to demonstrate the feasibility of our integrated system pipeline, rather than providing definitive evidence of the Enhanced NAO's impact. The participant pool was necessarily small (eight participants), limiting the statistical power and generalizability of results. Nevertheless, we performed statistical analyses and reported effect sizes alongside significance tests, presenting results as transparently as possible and allowing readers to interpret the findings in context. Importantly, the large observed effects indicate clear promise and motivate follow-up studies with fully powered designs. For example, detecting a medium effect size in a within-subjects design comparing two robot conditions across two gendered populations would require approximately 36 participants, which we plan for future work.\looseness-1

A second limitation involved a technical inconsistency in one session, where a participant interacted with the Enhanced NAO using a male TTS voice instead of the female voice presented in all other sessions. This mismatch may have influenced the system's performance and the participant's perception of the robot’s gender, and notably, this was the only participant who preferred the NAO AI. The deviation could also have interacted with turn-balance patterns. However, it also revealed the flexibility of our pipeline to support multiple voices. We did not analyze male vs. female voice as an experimental factor in this pilot, but future work should explicitly test how voice characteristics influence conversational quality and user preference. Future experiments will also implement stricter quality control to ensure consistent conditions across participants.

Additionally, participants were not blinded to the experimental condition. To mitigate bias, we used a counter-balanced condition order and asked participants to evaluate the conversation quality rather than the robot itself. Future studies could incorporate blinded or partially blinded setups to minimize potential bias due to visual differences between conditions.

Forth, because this pilot emphasized system integration over conversational design, prompt engineering and pretesting were not fully optimized, occasionally producing overly long or interview-like responses. These refinements are straightforward to implement and will be incorporated in future studies with larger participant pools, standardized conditions, and improved dialogue strategies.

Finally, although the new multimodal camera system was successfully integrated, we did not formally evaluate its perceptual or interaction benefits in this pilot. Integration was the primary goal; follow-up work will characterize the contributions of these sensing modalities in depth. Taken together, these limitations reflect the preliminary nature of the study, but also underscore its success in validating the technical feasibility of our integrated framework and establishing the groundwork for more comprehensive evaluation.
\vspace{-0.05cm}\section{Conclusion and future work}\vspace{-0.15cm}
This work demonstrates a pathway for revitalizing legacy social robots by integrating modern sensing, speech, and dialogue capabilities, demonstrated with the NAO. Our system restored usability while delivering higher conversational quality and stronger user preferences, without added latency, compared to the NAO AI, highlighting the value of extending rather than retiring such platforms.

Future work will address the remaining technical challenges to further improve interaction quality. In particular, we will migrate our speech-to-text and text-generation pipelines from cloud services to local processing using compact, high-performance models such as Gemma 3n \cite{gemma_3n_2025} or Qwen 3 \cite{qwen3technicalreport}. Running these models locally is expected to further reduce latency, strengthen privacy, and increase system autonomy. We will also leverage multimodal visual perception to detect and localize users prior to speech, enabling natural dialogue initiation. In multi-party settings, fused visual and thermal data will support more accurate speaker separation by disambiguating overlapping voices and tracking active participants, further improving responsiveness. 

Ultimately, we envision extending this framework beyond the NAO to demonstrate the platform-agnostic strategy for sustaining unsupported social robots. We are already deploying the framework to other legacy platforms like the Baxter and the TurtleBot 2, showcasing the system's flexibility for humanoid and mobile robots. More broadly, this work highlights how thoughtful system integration can transform aging robots into modern research tools, extending their lifespan and ensuring they continue to advance human-robot interaction.

\section*{Acknowledgments}\vspace{-0.15cm}
The authors thank the Ohio Space Grant Consortium for their generous support of the first author.

{  \footnotesize
\bibliographystyle{IEEEtran}

\bibliography{refs}

@misc{chatzilygeroudis2014naohelmet, title={{NAO Backpack and Helmet}}, url={https://www.thingiverse.com/costashatz/designs}, journal={Thingiverse}, author={Chatzilygeroudis, Konstantinos}, year={2014}, month={May}}

@misc{AmazonPolly_2025, title={Amazon Polly}, url={https://aws.amazon.com/polly/}, journal={Ai Voice Generator and text-to-speech tool - amazon polly - AWS}}

@article{mattamalaNaoBackpack2017,
    title={The NAO Backpack: An Open-hardware Add-on for Fast Software Development with the NAO Robot},
    author={Mattamala, Mat\'ias and Olave, Gonzalo and Gonz\'alez, Clayder and Hasb\'un, Nicol\'as and Ruiz-del-Solar, Javier},
    year={2017},
    url={https://github.com/uchile-robotics/nao-backpack}
}

@misc{bono2024openaccessnaooan,
      title={Open Access NAO (OAN): a ROS2-based software framework for HRI applications with the NAO robot}, 
      author={Antonio Bono and Kenji Brameld and Luigi D'Alfonso and Giuseppe Fedele},
      year={2024},
      eprint={2403.13960},
      archivePrefix={arXiv},
      primaryClass={cs.RO},
      url={https://arxiv.org/abs/2403.13960}, 
}

@misc{gestrin2024NaoChat,
  author={Gestrin, Elliot},
  title={NAOChat},
  year={2024},
  publisher={GitHub},
  journal={GitHub repository},
  howpublished={\url{https://github.com/ElliotGestrin/NAOChat}}
}

@article{gemma_3n_2025,
    title={Gemma 3n},
    url={https://ai.google.dev/gemma/docs/gemma-3n},
    publisher={Google DeepMind},
    author={Gemma Team},
    year={2025}
}

@misc{qwen3technicalreport,
      title={{Qwen3 Technical Report}}, 
      author={{Qwen Team}},
      year={2025},
      eprint={2505.09388},
      archivePrefix={arXiv},
      primaryClass={cs.CL},
      url={https://arxiv.org/abs/2505.09388}, 
}

@misc{respeaker_2023, 
    title={{ReSpeaker 4 Mic Array v2.0}}, 
    author={Seeed Studio},
    url={https://wiki.seeedstudio.com/ReSpeaker_Mic_Array_v2.0/},
    year={2023}
}

@misc{seek_2024, 
    title={{S314SPX}}, 
    author={{Seek Thermal}},
    url={https://shop.thermal.com/Mosaic-Core-Starter-Kit-320x240-57HFOV-FF}, 
    year={2025}
}

@misc{orbbec_gemini,
    title={{Gemini 2~L}}, 
    author={{Orbbec}},
    url={https://www.orbbec.com/products/stereo-vision-camera/gemini-2l/}, 
}

@misc{robotlab_2022,
    title={{NAO AI Edition}}, 
    author={{RobotLab}},
    url={https://www.robotlab.com/store/nao-ai-edition}, 
}

@misc{aldebaran_nao_2025,
    title={{NAO v6}}, 
    author={{Aldebaran}},
    url={https://aldebaran.com/en/nao6/}, 
}

@misc{aldebaran_altexttospeech_2025,
    author={{Aldebaran}},
    title={{ALTextToSpeech -- NAOqi 2.8 Documentation}},
    url={http://doc.aldebaran.com/2-8/naoqi/audio/altexttospeech.html#altexttospeech},
}

@misc{aldebaran_asr,
    title={{ALSpeechRecognition -- NAOqi 2.8 Documentation}},
    author={{Aldebaran}},
    url={http://doc.aldebaran.com/2-8/naoqi/audio/alspeechrecognition.html},
}

@misc{nao_image_2025,
    title={{NAOImage}}, 
    author={{NaoDevils}},
    url={https://github.com/NaoDevils/NaoImage}, 
}

@article{openai2024gpt,
  title={GPT-4o System Card},
  author={OpenAI},
  journal={arXiv preprint arXiv:2410.21276},
  year={2024}
}

@article{strathearnSobolewska2025nao, 
    title={Universities face getting stuck with thousands of obsolete robots – here’s how to avoid a research calamity}, 
    url={http://theconversation.com/universities-face-getting-stuck-with-thousands-of-obsolete-robots-heres-how-to-avoid-a-research-calamity-256829},
    journal={The Conversation}, 
    author={Strathearn, Carl and Sobolewska, Emilia}, 
    year={2025}, 
    month={May}
}

@article{willow_garage, 
    title={How a billionaire who wrote Google’s original code created a robot revolution}, 
    url={https://www.businessinsider.com/a-look-back-at-willow-garage-2016-2},
    journal={Business Insider}, 
    author={Jillian D'Onfro}, 
    year={2016}, 
    month={February}
}

@article{lawrenceCarol2019rethink, 
    title={Rise and Fall of Rethink Robotics}, 
    url={https://www.asme.org/topics-resources/content/rise-fall-of-rethink-robotics},
    journal={The Conversation}, 
    author={Lawrence, Carol}, 
    year={2019}, 
    month={April}
}

@article{nussey2021softbank, 
    title={SoftBank shrinks robotics business, stops Pepper production- sources}, 
    url={https://www.reuters.com/technology/exclusive-softbank-shrinks-robotics-business-stops-pepper-production-sources-2021-06-28/},
    journal={Reuters}, 
    author={Nussey, Sam}, 
    year={2021}, 
    month={June}
}

@article{deepgram2025transcription, 
    title={Deepgram Real-time Speech-to-Text}, 
    url={https://deepgram.com/product/speech-to-text},
    journal={Deepgram}, 
    author={Deepgram}, 
    year={2025}
}

@article{miller2025evaluating,
  title={Evaluating llm metrics through real-world capabilities},
  author={Miller, Justin K and Tang, Wenjia},
  journal={arXiv preprint arXiv:2505.08253},
  year={2025}
}

@article{rpi5_2025,
    title={Raspberry Pi 5},
    url={https://www.raspberrypi.com/products/raspberry-pi-5/},
    year={2025},
    journal={Raspberry Pi}, 
    author={Raspberry Pi}
}

@article{hallgren2012computing,
  title={Computing inter-rater reliability for observational data: an overview and tutorial},
  author={Hallgren, Kevin A},
  journal={Tutorials in quantitative methods for psychology},
  volume={8},
  number={1},
  pages={23},
  year={2012}
}

@inproceedings{fitter2016equipping,
  title={Equipping the Baxter robot with human-inspired hand-clapping skills},
  author={Fitter, Naomi T and Kuchenbecker, Katherine J},
  booktitle={2016 25th IEEE International Symposium on Robot and Human Interactive Communication (RO-MAN)},
  pages={105--112},
  year={2016},
  organization={IEEE}
}

@article{hailo20258l,
    author={Hailo Technologies LTD},
    title={Hailo-8L Entry-Level
AI Accelerator},
    journal={Hailo Technologies LTD},
    year={2025}
}

@article{grondin2018lightweightoptimizedsoundsource,
  title={Lightweight and Optimized Sound Source Localization and Tracking Methods for Open and Closed Microphone Array Configurations}, 
  author={Francois Grondin and Francois Michaud},
  year={2018},
  journal={arXiv preprint arXiv:1812.00115}
}

@misc{rethink2012baxter,
  title={Baxter Robot},
  author={Robotics, Rethink},
  year={2012}
}

@article{kumar2024kinematics,
  title={Kinematics \& Dynamics Library for Baxter Arm},
  author={Kumar, Akshay and Sahasrabudhe, Ashwin and Perugu, Chaitanya and Nirgude, Sanjuksha and Murugan, Aakash},
  journal={arXiv preprint arXiv:2409.00867},
  year={2024}
}

@mastersthesis{da2019control,
  title={The Control of Baxter Robot, and its Interaction with Objects Using Force Sensitive Ar10 Hands, Guided by Kinect.},
  author={da Cruz Lino, Pedro Jorge},
  year={2019},
  school={Universidade de Coimbra (Portugal)}
}

@inproceedings{devine2016real,
  title={Real time robotic arm control using hand gestures with multiple end effectors},
  author={Devine, Scott and Rafferty, Karen and Ferguson, Stuart},
  booktitle={International Conference on Control (CONTROL)},
  pages={1--5},
  year={2016},
  organization={IEEE}
}

@misc{kermorgant2025baxterros2,
  author={Kermorgant, Olivier},
  title={ROS2 Bridge for Baxter},
  year={2025},
  publisher={GitHub},
  journal={GitHub repository},
  howpublished={\url{https://github.com/CentraleNantesRobotics/baxter_common_ros2}}
}

@inproceedings{avalos2017real,
  title={Real-time teleoperation with the Baxter robot and the Kinect sensor},
  author={Avalos, Jose and Ramos, Oscar E},
  booktitle={IEEE Colombian conference on automatic control (CCAC)},
  pages={1--4},
  year={2017},
  organization={IEEE}
}

@misc{garage2012pr2,
  title={PR2 Robot},
  author={Garage, Willow},
  year={2012}
}

@article{chu2015robotic,
  title={Robotic learning of haptic adjectives through physical interaction},
  author={Chu, Vivian and McMahon, Ian and Riano, Lorenzo and McDonald, Craig G and He, Qin and Perez-Tejada, Jorge Martinez and Arrigo, Michael and Darrell, Trevor and Kuchenbecker, Katherine J},
  journal={Robotics and Autonomous Systems},
  volume={63},
  pages={279--292},
  year={2015},
  publisher={Elsevier}
}

@article{block2019softness,
  title={Softness, warmth, and responsiveness improve robot hugs},
  author={Block, Alexis E and Kuchenbecker, Katherine J},
  journal={International Journal of Social Robotics},
  volume={11},
  number={1},
  pages={49--64},
  year={2019},
  publisher={Springer}
}

@phdthesis{block2017should,
  title={How should robots hug?},
  author={Block, Alexis E},
  year={2017},
  school={University of Pennsylvania}
}

@article{block2023arms,
  title={In the arms of a robot: Designing autonomous hugging robots with intra-hug gestures},
  author={Block, Alexis E and Seifi, Hasti and Hilliges, Otmar and Gassert, Roger and Kuchenbecker, Katherine J},
  journal={ACM Transactions on Human-Robot Interaction},
  volume={12},
  number={2},
  pages={1--49},
  year={2023},
  publisher={ACM New York, NY}
}

@inproceedings{block2018emotionally,
  title={Emotionally supporting humans through robot hugs},
  author={Block, Alexis E and Kuchenbecker, Katherine J},
  booktitle={Companion of the 2018 ACM/IEEE International Conference on Human-Robot Interaction},
  pages={293--294},
  year={2018}
}

@article{wang2016pyride,
  title={PyRIDE: An Interactive Development Environment for PR2 Robot},
  author={Wang, Xun and Williams, Mary-Anne},
  journal={arXiv preprint arXiv:1605.09089},
  year={2016}
}

@inproceedings{ramezani2015smooth,
  title={Smooth robot motion with an optimal redundancy resolution for pr2 robot based on an analytic inverse kinematic solution},
  author={Ramezani, Nima and Williams, Mary-Anne},
  booktitle={IEEE-RAS International Conference on Humanoid Robots (Humanoids)},
  pages={338--345},
  year={2015},
  organization={IEEE}
}

@incollection{paul2021object,
  title={Object detection and pose estimation from rgb and depth data for real-time, adaptive robotic grasping},
  author={Paul, Shuvo Kumar and Chowdhury, Muhammed Tawfiq and Nicolescu, Mircea and Nicolescu, Monica and Feil-Seifer, David},
  booktitle={Advances in Computer Vision and Computational Biology: Proceedings from IPCV'20, HIMS'20, BIOCOMP'20, and BIOENG'20},
  pages={121--142},
  year={2021},
  publisher={Springer}
}

@inproceedings{song2019communication,
  title={Communication Effiency and User Experience Analysis of Visual and Audio Feedback Cues in Human and Service Robot Voice Interaction Cycle},
  author={Song, Haonan and Tan, JTC and Xing, Yi and Hou, Guan},
  booktitle={IEEE WRC Symposium on Advanced Robotics and Automation (WRC SARA)},
  pages={215--221},
  year={2019},
  organization={IEEE}
}

@article{alas2016controlling,
  title={Controlling a TurtleBot 2 through a web interface},
  author={Alas Escobar, Jorge and Holm, Anders},
  year={2016}
}

@inproceedings{bergeon2019raspberry,
  title={Raspberry Pi as an Interface for a Hardware Abstraction Layer: Structure of Software and Extension of the Turtlebot 2--Kobuki Protocol},
  author={Bergeon, Yves and K{\v{r}}iv{\'a}nek, V{\'a}clav and Motsch, Jean},
  booktitle={International Conference on Military Technologies (ICMT)},
  pages={1--6},
  year={2019},
  organization={IEEE}
}

@inproceedings{macias2023map,
  title={Map based localization using an RGB-D camera and a 2D LiDAR for autonomous mobile robot navigation},
  author={Macias, Luis Rodolfo and Aleman-Gallegos, J Enrique and Orozco-Rosas, Ulises and Picos, Kenia},
  booktitle={Optics and Photonics for Information Processing XVII},
  volume={12673},
  pages={112--122},
  year={2023},
  organization={SPIE}
}

@misc{garage2011turtlebots,
  title={TurtleBot 1 and 2},
  author={Garage, Willow},
  year={2011}
}

@article{rossi2022using,
  title={Using the social robot NAO for emotional support to children at a pediatric emergency department: randomized clinical trial},
  author={Rossi, Silvia and Santini, Silvano Junior and Di Genova, Daniela and Maggi, Gianpaolo and Verrotti, Alberto and Farello, Giovanni and Romualdi, Roberta and Alisi, Anna and Tozzi, Alberto Eugenio and Balsano, Clara},
  journal={Journal of Medical Internet Research},
  volume={24},
  number={1},
  pages={e29656},
  year={2022},
  publisher={JMIR Publications Toronto, Canada}
}

@article{amirova202110,
  title={10 years of human-nao interaction research: A scoping review},
  author={Amirova, Aida and Rakhymbayeva, Nazerke and Yadollahi, Elmira and Sandygulova, Anara and Johal, Wafa},
  journal={Frontiers in Robotics and AI},
  volume={8},
  pages={744526},
  year={2021},
  publisher={Frontiers Media SA}
}

@article{shamsuddin2012humanoid,
  title={Humanoid robot NAO interacting with autistic children of moderately impaired intelligence to augment communication skills},
  author={Shamsuddin, Syamimi and Yussof, Hanafiah and Ismail, Luthffi Idzhar and Mohamed, Salina and Hanapiah, Fazah Akhtar and Zahari, Nur Ismarrubie},
  journal={Procedia Engineering},
  volume={41},
  pages={1533--1538},
  year={2012},
  publisher={Elsevier}
}

@phdthesis{block2021huggiebot,
  title={HuggieBot: An interactive hugging robot with visual and haptic perception},
  author={Block, Alexis Emily},
  year={2021},
  school={ETH Zurich}
}

@inproceedings{block2021six,
  title={The six hug commandments: Design and evaluation of a human-sized hugging robot with visual and haptic perception},
  author={Block, Alexis E and Christen, Sammy and Gassert, Roger and Hilliges, Otmar and Kuchenbecker, Katherine J},
  booktitle={Proceedings of the 2021 ACM/IEEE international conference on human-robot interaction},
  pages={380--388},
  year={2021}
}
}

\end{document}